\documentclass[conference]{IEEEtran}

\usepackage[
    letterpaper,
    top=54pt,
    bottom=54pt,
    left=54pt,
    right=54pt
]{geometry}

\usepackage[english]{babel}

\usepackage{amsmath}
\usepackage{graphicx}
\usepackage[colorlinks=true, allcolors=blue]{hyperref}

\usepackage{url}
\usepackage{capt-of}
\usepackage{tabularx}
\usepackage{booktabs}
\usepackage{multirow}
\usepackage{comment}

\IfFileExists{latexml.sty}{
    \usepackage{latexml}
}{
    \newif\iflatexml
    \latexmlfalse
}

\newcommand{\figonecaption}{
SpectRobot framework.
(a) Example of a $224 \times 224$ grayscale input spectrogram.
(b) Learning pipeline, adapted from \protect\cite{TonyZhao2023}.
(c) Sorting task: the robot picks up and shakes an opaque box,
uses visual observations from top and wrist cameras, as well as tactile
spectrograms to infer one of four hidden contents, and places the box
in the corresponding bin.
(d) Box contents used in the sorting task. Scale bar indicates 2~cm.
}

\title{Learning tactile perception from high-bandwidth single-point sensing}

\author{
Joseph Rigal\textsuperscript{1}, Emmanuel Virot\textsuperscript{1,*},
Caroline Pascal\textsuperscript{2}
}

\IEEEoverridecommandlockouts

\iflatexml
\else
    \IEEEaftertitletext{
        \begin{center}
            \includegraphics[width=\textwidth]{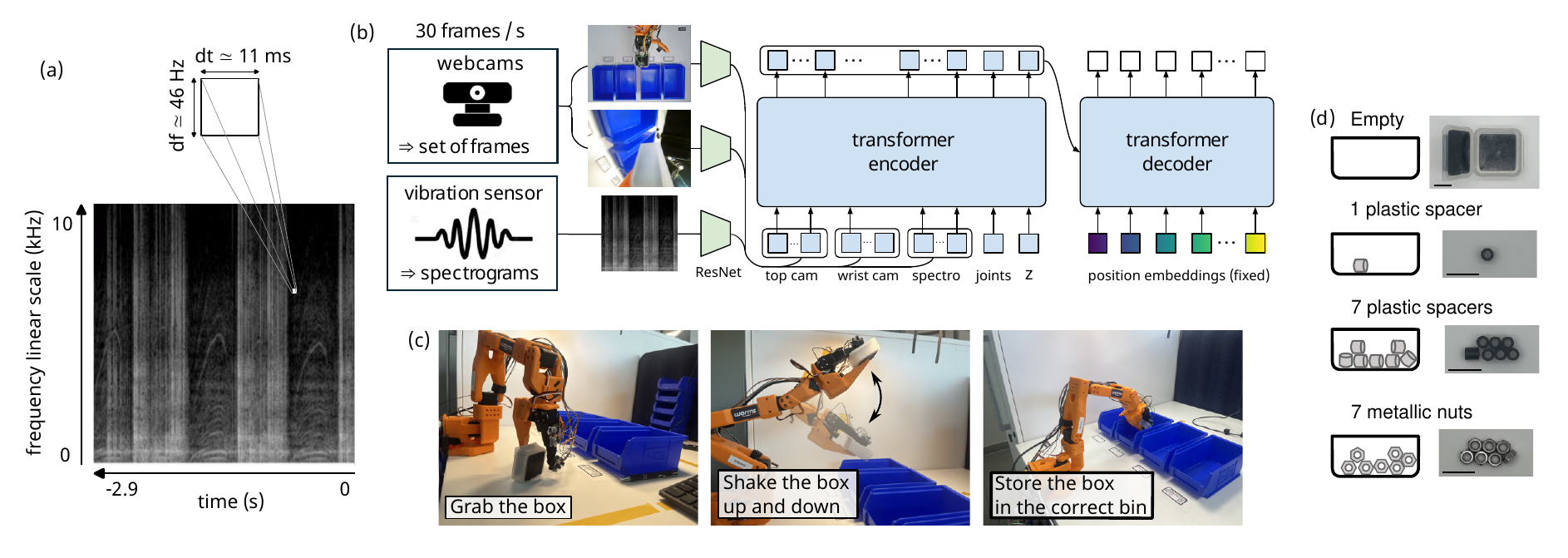}
            \captionof{figure}{\figonecaption}
            \label{fig:fig_01}
        \end{center}
    }
\fi

\begin{document}
\maketitle

\begingroup
\renewcommand{\thefootnote}{\arabic{footnote}}
\footnotetext[1]{Wormsensing, 2 Rue des Murailles, 38170 Seyssinet-Pariset, France.}
\footnotetext[2]{Hugging Face, 9 rue des Colonnes, 75002 Paris, France.}
\endgroup

\begingroup
\renewcommand{\thefootnote}{\fnsymbol{footnote}}
\footnotetext[1]{Corresponding author: \url{emmanuel.virot@wormsensing.com}.}
\endgroup

\begin{abstract}
Tactile sensing is increasingly being incorporated into learning-based robotic manipulation, yet many existing approaches rely on spatially distributed sensors.
Here we introduce {SpectRobot}, a framework that transforms single-point tactile signals into compact time-frequency spectrograms. 
These spectrograms encode high-bandwidth tactile histories as fixed-size image-like representations. 
They can be processed by standard vision encoders and integrated into learning pipelines originally developed for vision, while preserving temporal and frequency information unavailable to conventional cameras.

Rather than increasing spatial density through arrays of tactile elements, SpectRobot exploits the rich dynamics contained in sparse, high-bandwidth single-point measurements.
In our implementation, the sensors are mounted away from the contact surface while remaining mechanically coupled to it, reducing direct exposure to wear and potentially improving robustness in harsh environments and for long-term deployment on dexterous robots.

Our experiments demonstrate that:
(1) a robot can exploit single-point vibration signals to solve a visually occluded manipulation task; (2) temporal history strongly influences policy performance, while sensing bandwidth controls the spectral information available, with measurements extending to 100~kHz;
and (3) the same representation can be used across different tactile sensing technologies mediated by acceleration, force, or strain. 
We further show that capabilities previously associated with research-grade instrumentation can be accessed using readily available, off-the-shelf hardware.
We believe that broader access to high-bandwidth tactile sensing could facilitate the integration of contact dynamics into embodied learning systems and, for some tasks, offer an alternative or complement to increasing the spatial density of tactile sensing.
Our project page with videos is at \url{https://spectrobot-project.github.io/}.
\end{abstract}

\iflatexml
\begin{figure*}[t]
    \centering
    \includegraphics[width=\textwidth]{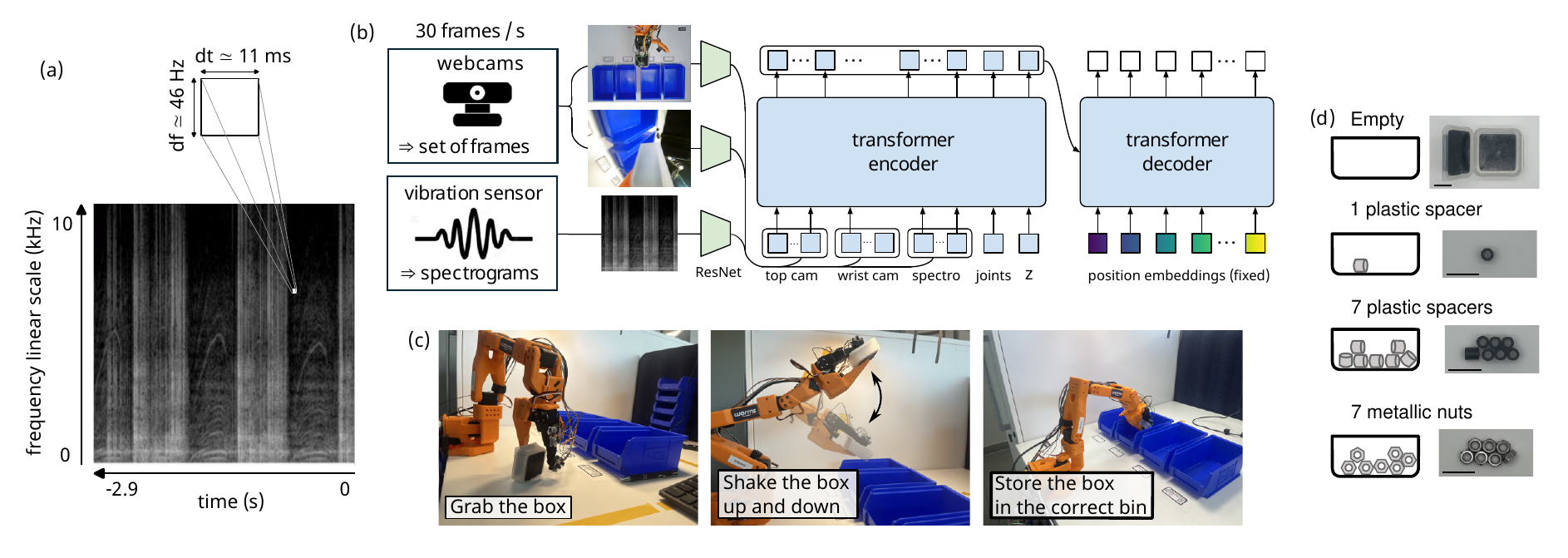}
    \caption{\figonecaption}
    \label{fig:fig_01}
\end{figure*}
\fi

\section{Introduction}

Does a robot need touch to learn dexterous manipulation? 
In one of the early demonstrations of learning-based robotic dexterity, OpenAI observed with surprise that ``tactile sensing is not necessary to manipulate real-world objects'' \cite{OpenAI2018}. 
This result highlighted the remarkable capabilities of vision-based policies, particularly when large amounts of simulated experience can be transferred to the physical world (sim-to-real approaches). Yet it also raises a broader question: is touch fundamentally unnecessary for robotic intelligence, or have current approaches learned to compensate for its absence?

Real-world interaction is inherently multimodal. 
Many physical processes relevant to manipulation are poorly observable by cameras, occluded by the gripper, or entirely invisible.
Slip, friction, surface interactions, and impacts are governed by local contact dynamics that can occur at timescales far beyond conventional visual acquisition. 
Slip detection in particular has motivated a broad range of tactile sensing approaches \cite{Romeo2020}. 
These interactions can also be difficult to reproduce faithfully in simulation. 
Contact mechanics combine friction, nonlinear elasticity, and high-speed structural dynamics, rapidly increasing the complexity of physical models and limiting sim-to-real transfer when the relevant interaction cannot be represented accurately. 
Direct tactile sensing therefore provides information that is complementary to vision rather than simply redundant with it.

The importance of this information is intuitive from human manipulation. 
Tasks such as buttoning a shirt, tying shoelaces, or adjusting the grip on a delicate object become considerably harder when tactile perception is impaired. 
Classical experiments on precision grip showed that humans continuously adapt grip force to frictional conditions using tactile information \cite{Westling1984}. 
Nevertheless, compared with vision, touch remains much less systematically integrated into modern learning-based robotic systems.

One reason is that there is no universal equivalent of the camera for touch.
Tactile information can be acquired through many physical modalities, such as electromagnetic effects, acceleration, strain, force, or pressure. 
Examples include motor currents \cite{Fajardo2021,Zhuo2016}, piezoelectric vibration sensors \cite{Ayral2023}, resistive sensors \cite{Koiva2013,Tao2026,Huang_Binghao2026}, contact microphones \cite{GamboaMontero2020,Du2022,Liu2024,JialiangZhao2025}, magnetic skins \cite{Pattabiraman2024,Pattabiraman2025,ZifanZhao2025}, and optical deformation sensors \cite{Ford2023,Vitrani2025}. These sensors measure different projections of the interaction, with different sensitivities, bandwidths, spatial resolutions, integration constraints, and costs. 
Recent efforts to systematically compare tactile sensing technologies further illustrate the diversity of the available design space \cite{Chung2026,Zorin2026}. 
Consequently, adding touch to an existing robot can require substantial changes in the mechanical structure of the gripper, acquisition electronics, calibration, signal processing, and learning pipelines.

A second challenge is how tactile information should be represented for learning. 
Modern robotic learning has inherited much of its architecture from computer vision. 
Transformer-based manipulation policies and multimodal learning systems typically treat camera images as their primary representation of the environment \cite{TonyZhao2023,OpenAI2018,Cadene2024}. 
When touch is added, a natural strategy is therefore to spatialize it, for example by arranging measurements from tactile arrays into matrices analogous to images \cite{Pattabiraman2024,Zhang2026}. 
This approach has enabled increasingly rich tactile perception, but it implicitly emphasizes spatial resolution as the principal route toward increasing tactile information.
However, we believe that touch cannot be fully represented by spatial information alone. 
Contact generates temporal dynamics whose characteristic frequencies can extend orders of magnitude beyond the acquisition rate of conventional cameras. 
A single sensing point can therefore contain substantial information when sampled with sufficient temporal resolution. 
Vibration-based sensing has already shown that time-frequency representations can reveal contact phenomena such as slip and manipulation dynamics \cite{Ayral2023}. 
Recent work has further shown that acoustic signals can resolve manipulation states that are ambiguous from vision alone, through active interaction and temporal audio or vibration context \cite{Liu2024,Mao2026,Sandykbayeva2022}. 
Here, we focus instead on the sensing and representation design space, specifically how acquisition bandwidth, temporal context, and sensor type affect learning. Rather than asking only how many tactile elements should cover a robot surface, this suggests a complementary question: how much information can be recovered by increasing the bandwidth of a small number of sensing points?

Time-frequency representations, Fig.~\ref{fig:fig_01}(a), provide a particularly convenient interface for addressing this question. 
A signal sampled at high rate can be transformed into a spectrogram in which one axis represents time, the other frequency, and pixel intensity represents power spectral density. Importantly, these spectrograms are image-like representations with explicit temporal and spectral coordinates. They can therefore be supplied directly to architectures and learning pipelines originally developed for vision, as shown in Fig.~\ref{fig:fig_01}{(b)}, while preserving physical information that is inaccessible to conventional cameras. 
Spectrogram-based representations have already enabled time-frequency information to be integrated into learning pipelines \cite{Ayral2023,Liu2024,Mejia2024,Li2022}. This approach also compresses long, high-rate temporal streams into fixed-size representations and allows the learning algorithm itself to identify which frequencies and temporal regions are useful for the task. 
Here we introduce SpectRobot, a framework that gives learning-based robots access to high-bandwidth tactile and vibrational information by representing physical interaction as time-frequency images. 
We demonstrate the added capability through a deliberately visually occluded manipulation task. 
Several opaque boxes contain different objects that cannot be distinguished from visual information alone. 
The robot interacts with one box by gently shaking it up and down (causing its contents to move), as shown in Fig.~\ref{fig:fig_01}{(c)}, and must infer its content, Fig.~\ref{fig:fig_01}{(d)}, from the resulting mechanical response. 
This setting provides a controlled example in which vision cannot directly access the physical variable required to solve the sorting task. 

This work presents four key contributions. 
(1) We introduce a computationally lightweight framework for converting high-bandwidth tactile and vibrational signals into image representations that can be processed directly by learning architectures originally developed for vision. 
(2) We demonstrate autonomous robotic manipulation using tactile information extending up to 100~kHz, providing access to contact dynamics far beyond the temporal resolution of conventional vision. 
The analysis of attention maps and scene sensitivity on the spectrograms allows us to inspect which regions of the time-frequency representation contribute to the learned behavior.
(3) We systematically investigate how sensing bandwidth and temporal history affect learned performance, allowing the useful temporal and spectral scales to emerge from the task rather than being prescribed beforehand. 
(4) We benchmark the paradigm across seven types of sensors {and two acquisition chains} spanning different levels of cost and measurement fidelity, from research-grade instrumentation to readily available, off-the-shelf hardware. 

Together, these results suggest that using high-bandwidth spectrograms from a single sensing point provides an alternative or complement to increasing spatial sensing density and offers a practical route for integrating rich physical interaction into modern embodied intelligence.

\section{Material and methods}

\subsection{Robotic platform and imitation-learning framework}

We implemented SpectRobot within the open-source LeRobot framework \cite{Cadene2024}, which provides a common infrastructure for dataset acquisition, multimodal observations, imitation learning, and deployment on physical robots, while preserving compatibility with existing learning pipelines and lowering the barrier to integrating high-bandwidth tactile sensing on custom hardware. 
Experiments were performed using an SO-101 robotic arm, Fig.~\ref{fig:fig_01}{(c)}, manufactured and pre-calibrated by WowRobo. 
The robot was operated using a leader-follower configuration during data acquisition. 

For policy learning, we used the Action Chunking with Transformers (ACT) architecture \cite{TonyZhao2023}, as detailed in Fig.~\ref{fig:fig_01}{(b)}. 
Visual observations were encoded using a ResNet-18 backbone whose weights were trained jointly with the transformer architecture.  
Tactile signals were converted into spectrogram images and supplied to the learning architecture using the same input representation as a conventional camera stream. 
Our emphasis on real-world data acquisition (instead of a sim-to-real approach) is deliberate. 
Simplifying complex interactions (friction, impacts, nonlinear deformation) for simulation may remove precisely the high-frequency physical information that SpectRobot is designed to exploit. We therefore train the policies directly from physical demonstrations. 

\subsection{High-bandwidth acquisition}

To investigate tactile information over a large frequency range, we used a Dewesoft IOLITE\textsuperscript{\textregistered}-X-8xACC  acquisition system running an openDAQ\textsuperscript{\texttrademark} server (open-source framework \cite{OpenDAQ2026}). The acquisition system provides 24-bit conversion, simultaneous sampling at up to 200~kS/s per channel, and signal-to-noise performance characterized through a preliminary benchmark.
This configuration allowed us to investigate tactile vibration content up to approximately 100~kHz (Nyquist limit).

The chosen bandwidth is intentionally much larger than the ranges typically considered in robotic tactile sensing and slip detection, which frequently concentrate on frequencies below 1~kHz \cite{Friedl2021,Meier2016,Romeo2020}. 
These ranges are often considered because they overlap with the frequency region of greatest human vibrotactile sensitivity in the range 100--400~Hz \cite{Verrillo1969,Handler2021}, but they exclude higher-frequency mechanical phenomena that can occur during contact in rigid robotic grippers made of plastic or metal.

Our resulting acquisition rate is also fundamentally different from that of conventional robotic vision. 
A camera operating at approximately 30 frames per second samples the environment every 33~ms, whereas the tactile acquisition system records at intervals of 5.0~$\mu$s, i.e., about 6,600 tactile samples per camera frame. 
SpectRobot does not directly feed this raw stream to the policy. 
Instead, successive windows of the signal are transformed into spectrograms, producing a compact representation that can be delivered to the learning architecture at a rate comparable to the visual stream.
Spectrograms are generated during training-dataset acquisition, and raw time-domain signals are not retained. 
Consequently, our approach can scale to higher tactile sampling rates without increasing the size or rate of the policy input, and therefore without directly increasing the computational cost of policy training or inference. 
Higher sampling rates may, however, increase acquisition and spectrogram-generation costs.

In the present implementation, camera frames and spectrograms were generated every 33~ms, while the robotic policy issued motor commands approximately every 33~ms as well.
The tactile acquisition bandwidth and the policy update frequency are therefore deliberately decoupled: high-frequency information is captured within each tactile observation even though the robot does not produce motor commands at the corresponding acquisition rate.
For the experiments reported here, spectrograms were computed as power spectral density estimates using a Tukey window, 50\% overlap, $n_\text{FFT}$ from 64 to 4096 samples, and sampling rates up to 200~kS/s, depending on experimental conditions.
Each spectrogram was converted to a 224 $\times$ 224 grayscale image before being supplied to the ResNet-18. 
Unlike RGB images, whose three-channel representation is rooted in human trichromatic vision, tactile sensors produce scalar signals, such that each spectrogram pixel is naturally represented by a single scalar value. 
We therefore used grayscale rather than introducing an arbitrary mapping to RGB channels. 
This representation was sufficient for the experiments reported here, and alternative image bit depths were not evaluated.
The spectrogram frequency bandwidth and temporal context were varied to assess their effects on policy learning.
Before the full benchmark, we tested four frequency ranges on a simplified two-box task: 0--1~kHz (classically considered in the literature), 5--6~kHz, 0--10~kHz, and 0--100~kHz.
The 0--1~kHz range yielded the lowest performance and appeared to be particularly affected by motor-related noise, while the 0--10~kHz and 0--100~kHz ranges showed similar performance, both outperforming the narrower bands.
We therefore selected 0--10~kHz as the reference condition.

It is worth noting that many acoustic machine-learning applications use Mel spectrograms, which warp the frequency axis according to human auditory perception, providing greater resolution at low frequencies and compressing higher frequencies.
While appropriate for speech and audible sound, this perceptual bias has no obvious basis for robotic tactile sensing. SpectRobot therefore uses a linear frequency scale, which preserves the natural frequency spacing of harmonics in mechanical vibration signals.

\subsection{Sensors evaluated}
\begin{figure*}[ht]
\centering
\includegraphics[width=1\linewidth]{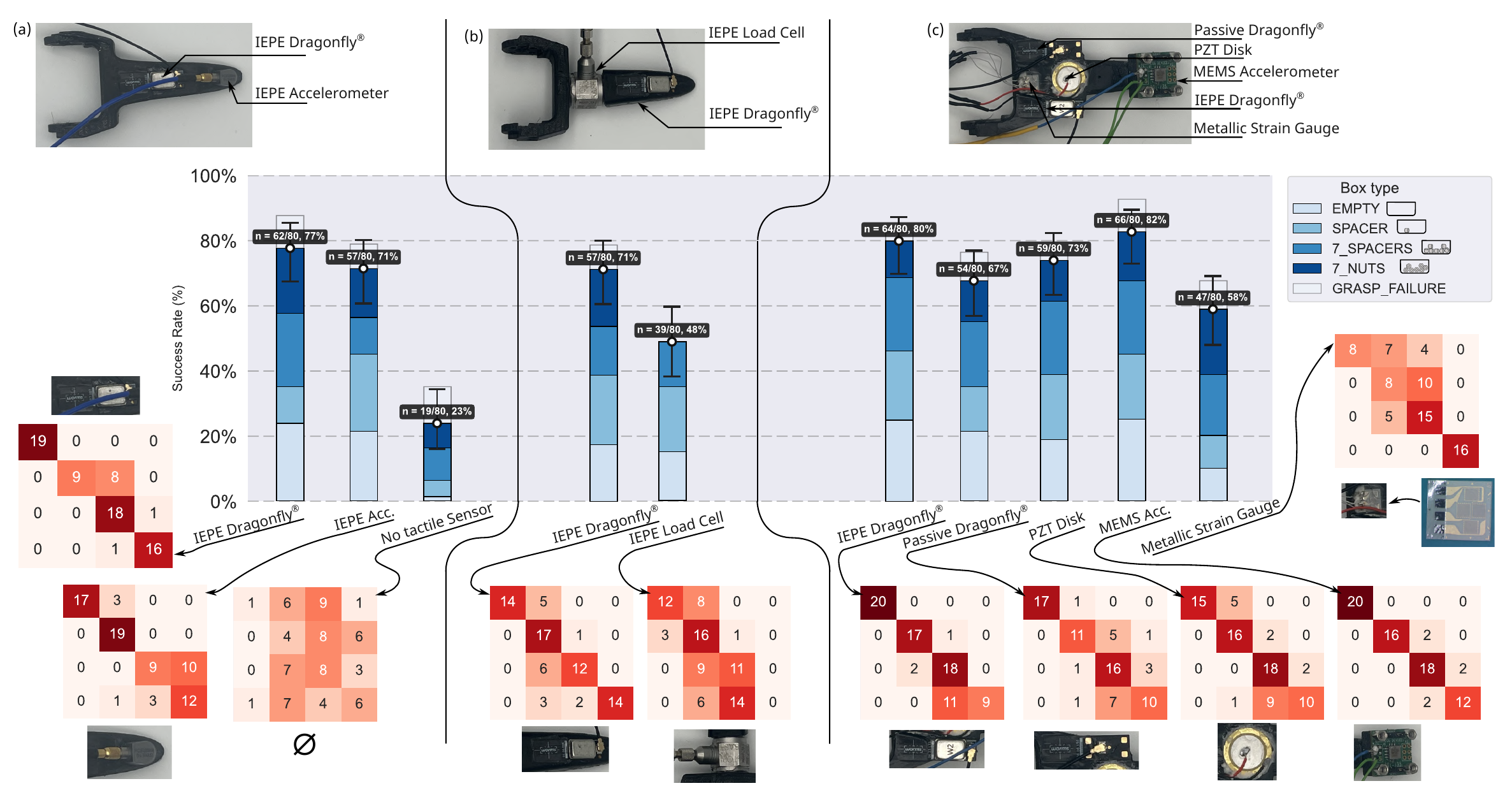}
\caption{\label{fig:fig_02} 
Results of the autonomous rollouts. 
(a,b,c) The three gripper designs. 
(middle) Success rates with 95\% Wilson score confidence intervals, calculated following \cite{Capuano2026}. 
(bottom) Confusion matrices for each sensor.
Rows indicate the actual box content (empty, 1 spacer, 7 spacers or 7 metallic nuts), while columns indicate the blue bin selected by the robot.
}
\end{figure*}

All the sensors evaluated in this study, Table~\ref{tab:tab_01}, were integrated on the upper part of the gripper, far from the actual contact points, as shown in Fig.~\ref{fig:fig_02}. 
They remained mechanically coupled to the contact points while being physically separated from direct abrasion as can occur with soft tactile skins and pressure arrays \cite{Lambeta2020,Bhirangi2024}.

Some of the piezoelectric sensors evaluated in this study incorporate built-in impedance conversion electronics, commonly referred to as Integrated Electronics Piezo-Electric (IEPE).
This reduces output impedance and limits parasitic effects associated with electromagnetic interference and triboelectric cable noise. IEPE is a standard that is particularly relevant in robotic systems, where cable motion, actuator vibrations, electromagnetic interference, and structural resonances can contaminate low-amplitude measurements. 
We selected IEPE sensors covering three distinct physical modalities of touch: force, strain, and acceleration. 
Piezoelectric sensors are particularly well suited to SpectRobot because they are highly sensitive to transient and vibrational phenomena over a broad bandwidth. 
However, they measure dynamic signals rather than true static contact quantities and therefore do not directly provide quantities such as the static friction coefficient or normal contact force. 

\newcommand{\costdot}{\mathord{\bullet}}
\newcommand{\costdots}[1]{\ifcase#1\or
    \costdot\or
    \costdot\mkern2mu\costdot\or
    \costdot\mkern2mu\costdot\mkern2mu\costdot\or
    \costdot\mkern2mu\costdot\mkern2mu\costdot\mkern2mu\costdot
  \fi
}

\begin{table}[ht]
\centering
\caption{Sensing technologies evaluated with SpectRobot.}
\label{tab:tab_01}

\footnotesize
\setlength{\tabcolsep}{3pt}

\begin{tabularx}{\columnwidth}{@{}X l l c@{}}
\toprule
\textbf{Sensor} &
\textbf{Reference} &
\textbf{Measurement} &
\textbf{Cost} \\
\midrule

MEMS accelerometer
& ADXL356CZ
& Acceleration [$g$]
& $\costdots{1}$ \\

IEPE accelerometer
& PCB P. TLD352A56
& Acceleration [$g$]
& $\costdots{4}$ \\

IEPE load cell
& PCB P. 208C02
& Force [N]
& $\costdots{3}$ \\

Metallic strain gauge
& HBK VA73K1.6/350\_E
& Strain [$\mu$m/m]
& $\costdots{1}$ \\

IEPE Dragonfly\textsuperscript{\textregistered}
& DGF-UNI-W220405-10
& Strain [$\mu$m/m]
& $\costdots{2}$ \\

Passive Dragonfly\textsuperscript{\textregistered}
& DGF-UNI-AA20405-10
& Strain [$\mu$m/m]
& $\costdots{2}$ \\

PZT disk
& Ø~15~mm
& Voltage [V]
& $\costdots{1}$ \\

\bottomrule
\end{tabularx}

\vspace{2pt}

\scriptsize
Cost categories:
$\costdots{1}$ $<$100~€;
$\costdots{2}$ 100--500~€;
$\costdots{3}$ 500--1000~€;
$\costdots{4}$ $>$1000~€.

\end{table}

\textbf{MEMS accelerometers} provide frequency response down to static acceleration and can therefore capture gravity, slow gripper motion, impacts and structural vibration within a compact and inexpensive device. 
Commonly integrated into inertial measurement units (IMUs), MEMS accelerometers typically rely on a microscopic proof mass suspended by compliant structures, whose acceleration-induced displacement is measured, often capacitively. 
Their mechanical resonance and readout electronics limit the usable bandwidth, typically to a few kilohertz for general-purpose devices. 

\textbf{IEPE accelerometers} instead emphasize dynamic structural excitation.
They are based on piezoelectricity and cannot measure true static acceleration. 
Their stiff sensing element, high resonance frequency and IEPE conditioning make them well suited to low-noise, high-bandwidth vibration measurements. 

\textbf{IEPE load cells} are also based on piezoelectricity and provide a measurement of the resultant force transmitted through the gripper structure, with high sensitivity and broad dynamic bandwidth. The main limitations are the absence of true static response and the fact that they measure the force transmitted at their mounting location rather than local contact deformation. 
To place the load cell in the load path, the gripper had to be redesigned, as shown in Fig.~\ref{fig:fig_02}.
While load cells have historically been widely used in robotics and are often based on resistive strain gauges, the IEPE load cell evaluated here relies on piezoelectricity.

\textbf{Metallic strain gauges} measure local deformation through changes in electrical resistance and can capture static strain. 
They require a Wheatstone bridge, and in this study we used a full-bridge configuration to compensate temperature effects and minimize measurement noise. 
Their main disadvantages are complex installation and soldering, low electrical output, sensitivity to temperature, bonding and alignment requirements, and the need for stable bridge excitation and amplification.

\textbf{IEPE and passive Dragonfly\textsuperscript{\textregistered}} sensors measure strain using a $10~\mu$m-thick unidirectional piezoelectric layer with an active area of a few square millimeters. 
Their small mechanical footprint, low mass, high sensitivity, and broad dynamic response make them well suited for detecting weak structural events without requiring a dedicated compliant load path. 
The IEPE version has a gain of $\times 10$, a specified bandwidth extending from 0.04~Hz to 100,000~Hz, and provides a low-impedance output suitable for long cables and electrically noisy robotic environments. 
With the present acquisition chain, including ambient office noise, the system achieves a dynamic strain resolution of 0.06~$\mu$m/m, corresponding to interactions of approximately 0.3~mN at the gripper of the SO-101 robot (light-contact detection).
This type of sensor was bonded to each gripper design, Fig.~\ref{fig:fig_02}, as a common reference channel, allowing acquisition consistency and dataset-level shifts to be monitored before comparing to other sensors. The passive Dragonfly\textsuperscript{\textregistered} relies on the same piezoelectric principle but outputs charge directly, requiring no sensor-side power and allowing greater flexibility in the choice of acquisition chain.

\textbf{Bare lead zirconate titanate (PZT)} was included as a low-cost implementation of piezoelectric structural sensing \cite{Mao2026}. 
PZT offers high piezoelectric response and can effectively capture impacts, vibration and slip-related transients. 
Its low component cost is a major advantage. 
However, bulk PZT suffers from brittleness and environmental concerns associated with toxic lead oxide, particularly its volatilization during high-temperature processing, while its response is also highly sensitive to electrical connections, cabling, and shielding \cite{Panda2015,Chen2024}.

The benchmark was designed to determine whether the SpectRobot representation depends on a specific tactile technology or can instead provide a common interface for sensors measuring different physical quantities. 
Several alternative sensing approaches were not retained because of limited high-frequency bandwidth, integration complexity, or reduced robustness arising from effects such as wear, drift with temperature, hysteresis, and calibration variability. 
These approaches included motor currents, Polyvinylidene fluoride (PVDF), force-sensitive resistors (FSR), and piezoresistive, magnetic or capacitive tactile arrays, each offering practical advantages but also limitations for the high-bandwidth, robust sensing targeted here.

\subsection{Data collection and policy training}

Training demonstrations were acquired through leader-follower teleoperation under constant artificial lighting conditions. 
Each dataset contained {40} demonstration episodes. 
During acquisition, we sought to avoid introducing trivial visual correlations between the training and evaluation conditions.  

Policies were trained on a consumer-grade Ubuntu workstation equipped with an Intel Core i5 processor, 32 GB of RAM, and an NVIDIA RTX 5070 GPU.
Training used LeRobot version 0.5.2 with 100,000 optimization steps per model. 
Under this configuration, a typical training run took approximately three hours.
For each experimental condition, 80 autonomous inference episodes were conducted, with 20 episodes per box type, Fig.~\ref{fig:fig_01}{(d)}, and the resulting success rates were computed.

Because the teleoperation framework was originally designed for vision-based demonstrations, the operator received no direct force feedback from the gripper. 
We explored real-time spectrogram visualization and audio playback as alternative forms of tactile feedback in separate preliminary tests. Neither form of feedback was used during the training demonstrations analyzed in this study.

\section{Results}

\subsection{Tactile sorting under visual occlusion}

The task is designed to test whether the policy can exploit tactile information to infer visually occluded object properties. 
It consists of a sorting task involving an opaque box placed on a table, with four visually indistinguishable classes: empty, one plastic spacer, seven plastic spacers, and seven metallic nuts, Fig.~\ref{fig:fig_01}{(d)}. 
During each inference episode, the robot picks up the box, gently shakes it, and receives both visual observations and the resulting tactile spectrogram. 
It then places the box into the corresponding blue bin, as shown in Fig.~\ref{fig:fig_01}{(c)}.
This sequence is representative of industrial pick-and-place and sorting operations in which objects must be classified and routed according to properties that cannot be inferred from their external appearance alone.

The four classes were deliberately chosen to make tactile discrimination non-trivial from the spectrograms alone.
The robot must distinguish both between different numbers of objects made of the same material and between different materials with the same number of objects.
Although the box content can be easily identified by hand while shaking it, distinguishing the classes from the corresponding spectrograms alone is considerably more difficult. This four-class setup was therefore designed to be substantially more challenging than a simpler empty-versus-filled classification considered in preliminary experiments.

\subsection{Sensor benchmark}

We first evaluated how the tactile modality affects task performance using spectrograms spanning 0--10~kHz with $n_{\text{FFT}}=512$~samples, corresponding to 2.9~s of temporal history, as shown in Fig.~\ref{fig:fig_02}. 
Success rates in the bar plots were computed over all inference episodes, including grasp failures. 
A grasp failure was defined as an episode in which the robot failed to grasp the box correctly or dropped it before reaching a blue bin. 
A trial was considered successful only if the box was placed in the bin corresponding to its content. 
The confusion matrices report only episodes in which the box was successfully delivered to one of the bins, so some rows contain fewer than the 20 inference episodes performed for each box type. 
This behavior is consistent with the tactile representation contributing not only to box classification but also to grasp control: replacing the tactile spectrogram with a blank image at inference time prevented the policy from completing the grasp, suggesting an entanglement between tactile and motor information rather than a purely additive classification signal. 
This observation was obtained from a single ablated model and remains preliminary.

The vision-only policy achieved 23\% success, close to the 25\% chance level, whereas all tactile conditions improved performance. 
Across the three grippers, the IEPE Dragonfly\textsuperscript{\textregistered} reference sensor achieved success rates of 77\%, 71\%, and 80\%, supporting the comparability of the independently recorded datasets. 
The other tactile sensors achieved 71\% for the IEPE accelerometer, 48\% for the IEPE load cell, 67\% for the passive Dragonfly\textsuperscript{\textregistered}, 73\% for the PZT disk, 82\% for the MEMS accelerometer, and 58\% for the metallic strain gauge. 
The 95\% Wilson score confidence intervals of several of the best-performing sensors overlap, so their relative ranking is not statistically resolved at the present sample size.

Despite the overlapping confidence intervals, a broad group of comparatively high-performing sensors emerges, including both accelerometers, both Dragonfly\textsuperscript{\textregistered} variants, and the PZT disk. 
These technologies span markedly different trade-offs in integration, cost, bandwidth, and robustness. 
They range from the very low-cost PZT disk to Dragonfly\textsuperscript{\textregistered} sensors with bandwidth extending up to 100~kHz, while the MEMS accelerometer is limited by a resonance around 5~kHz.
This diversity suggests that, for the sorting task considered here, strong performance is not restricted to a single sensing technology or hardware class.

The confusion matrices in Fig.~\ref{fig:fig_02} show that classification errors depend on the sensing technology. 
Rows indicate the actual box content, while columns indicate the blue bin selected by the robot.
Empty boxes are generally well identified by the best-performing sensors, while most errors occur between the filled classes. 
Several sensors show confusion between seven plastic spacers and seven metallic nuts, whereas the metallic strain gauge exhibits broader confusion among the plastic-content classes while clearly identifying the metallic nuts. 
Interestingly, the three IEPE Dragonfly\textsuperscript{\textregistered} confusion matrices also exhibit different classification errors despite similar overall success rates, suggesting that the detailed error structure is sensitive to the gripper configuration and independently recorded dataset.

We adopted the ACT imitation-learning architecture because it is relatively fast to train and, importantly, trains both the ResNet encoder and the transformer from scratch. 
This allows the entire policy to adapt directly to tactile spectrograms, rather than relying on a pretrained vision backbone to already interpret them as meaningful inputs. 
As with other imitation-learning approaches, performance can still vary with the coverage and quality of the training dataset, and human-in-the-loop data collection such as DAgger could reduce this dependency in future work.

\subsection{Effect of bandwidth and temporal context}

We next compared spectrogram configurations spanning different spectral bandwidths and temporal contexts using the IEPE Dragonfly\textsuperscript{\textregistered} sensor. 

Each spectrogram was resized and replicated identically across the RGB channels to match the 224$\times$224$\times$3 input shape expected by the ResNet-18 encoder. 
For a given bandwidth, this corresponds to a frequency spacing of $(\text{bandwidth})/224$ per pixel. 
With 50\% overlap between successive Fourier windows, the temporal spacing is $dt = n_{\text{FFT}}/(2f_s)$, giving a total temporal history of $224 \times n_{\text{FFT}}/(2f_s)$. 
For each bandwidth, $n_{\text{FFT}}$ was therefore selected from powers of two to obtain temporal histories close to 0.3~s and 3~s, referred to hereafter as the short-window and long-window configurations, respectively.
Therefore, frequency bandwidth and total duration were not independently controlled, and the experiments should be interpreted as comparisons between spectrogram configurations rather than isolated ablations of temporal context or frequency content.

Four spectrogram configurations were compared: 10~kHz with temporal windows of 0.36~s (short) and 2.9~s (long), and 100~kHz with temporal windows of 0.29~s (short) and 2.3~s (long). For the short-window observations, the success rate remained close to chance, reaching approximately 31\% at 10~kHz and 32\% at 100~kHz. 
With the longer temporal windows, the success rate increased to approximately 86\% at 10~kHz and 92\% at 100~kHz. 
Across these configurations, performance was substantially higher for the longer-context spectrograms than for the shorter-context spectrograms.

For the long-window configurations, which had comparable temporal durations, the 0--100~kHz condition yielded a slightly higher success rate than the 0--10~kHz condition, although their 95\% Wilson score confidence intervals overlap and the difference is not statistically resolved with the present number of inference episodes. 

For the present sorting task, the additional information above 10~kHz therefore appears to provide limited benefit compared with providing sufficient temporal context. 
This does not imply that frequencies above 10~kHz are generally unnecessary. 
Tasks dominated by higher-frequency contact dynamics, such as frictional interactions, abrasive material removal, or ultrasonic dental and surgical tools (tested in preliminary experiments), may instead benefit from sensing bandwidths extending toward 100~kHz.

\subsection{Attention and scene-sensitivity analysis}

To investigate which parts of the tactile representation are used by the learned policy, we analyzed the trained ACT models using Emboviz, a visualization framework developed for LeRobot \cite{Emboviz2026}. 
We considered two complementary diagnostics. 
Attention maps report internal attention weights within the model's forward pass. 
Scene sensitivity quantifies how the policy output changes when local regions of the input image are occluded, producing a spatial saliency map. 
To obtain more robust trends, we averaged the maps over time to retain only their frequency dependence, and then averaged across successful episodes within each box class to reduce random fluctuations.
These diagnostics show which parts of the input the model focuses on or is sensitive to, but they do not explain why the policy makes its decisions.
Such information could ultimately help guide the design of task-specific indicators for complex fine-manipulation operations.

For the IEPE Dragonfly\textsuperscript{\textregistered} policy using 0--10~kHz spectrograms with 2.9~s of temporal history, attention increased markedly during shaking, particularly for the seven-plastic-spacer and seven-metallic-nut classes. 
For non-empty boxes, attention was strongest between 2~kHz and 4~kHz near the end of the shaking motion.
Scene sensitivity showed an even broader increase during shaking across most of the 0--10~kHz range, indicating that a large portion of the measured spectrum could influence the policy output.

\subsection{Low-cost, lower-fidelity acquisition setup}

Although the acquisition system used in the experiments above provides 24-bit digitization, the processing pipeline converts the spectrograms to 8-bit grayscale before they are supplied to the policy. 
We therefore evaluated a lower-cost acquisition setup using three of the best-performing sensors from the benchmark: the MEMS accelerometer, the IEPE Dragonfly\textsuperscript{\textregistered}, and the PZT disk. 
The setup consisted of an IEPE interface converter module (ZONRI), an ADS8688 analog-to-digital converter (Texas Instruments), and a Teensy 4.1 microcontroller (PJRC), for a total cost on the order of 100~€. 
The IEPE interface converter was used only with the IEPE Dragonfly\textsuperscript{\textregistered}, providing a low-impedance output with improved immunity to electromagnetic interference. 
The other sensors are not IEPE-compatible and therefore cannot benefit from the same conditioning approach.
The ADS8688 is a 16-bit, eight-channel successive-approximation-register (SAR) analog-to-digital converter (ADC) interfaced through a Serial Peripheral Interface (SPI). 
It provides bipolar input ranges, a maximum aggregate sampling rate of 500~kS/s, and an integrated second-order anti-aliasing low-pass filter with a typical $-3$~dB cutoff frequency of 15~kHz, making it well suited to the 0--10~kHz configuration used here. 
The Teensy 4.1 provides sufficient computational and timing margin for deterministic acquisition and real-time data handling.

The RMS noise of this low-cost acquisition chain was approximately one order of magnitude higher than that of the acquisition system used in the experiments reported above.
Nevertheless, this level of measurement fidelity was sufficient for the relatively energetic interaction considered here, in which the robot actively shakes the box.
In a separately collected but comparable dataset for the same four-class sorting task, the PZT disk, the IEPE Dragonfly\textsuperscript{\textregistered}, and the MEMS accelerometer achieved success rates of 93\%, 83\%, and 70\%, respectively, all well above the 42\% obtained without tactile sensing.
These results suggest that substantially lower-cost acquisition hardware can retain the information required by the learned policy when the mechanical signals are sufficiently large. 
For tasks involving subtler contacts or lower-amplitude interactions, however, the advantages of a higher-fidelity acquisition chain may become more important.
In particular, low-noise IEPE conditioning and high-sensitivity sensors capable of resolving sub-microstrain or nanostrain-scale signals may become necessary when the relevant tactile information approaches the measurement noise floor.

\section{Discussion and further directions}

As shown in the results above, our benchmark identified piezoelectric sensing technologies and the MEMS accelerometer as the best-performing sensors among those evaluated here.
Their dynamic nature also suggests complementary multimodal strategies. 
Piezoelectric sensors are particularly sensitive to transient and vibrational phenomena but do not directly capture static loading. 
Their measurements could therefore be combined with lower frequency information, such as force estimates obtained from actuator currents, to provide a more complete description of interaction ranging from static loading to high-frequency contact dynamics.

An important extension would be to combine the richness demonstrated here with spatial tactile information. 
Several vibration sensors distributed over a robotic structure could provide complementary spectrogram observations. 
These measurements could be treated as additional image streams, and encoded as additional channels, alongside conventional RGB or depth observations. With sufficiently synchronized sensors and appropriate placement, differences between their responses could potentially provide information about the localization and propagation of contact events.
Such an approach could bridge high-bandwidth sensing and spatial tactile arrays, combining spatial localization with rich time-frequency dynamics in a unified, largely sensor-agnostic pipeline.

Beyond the demonstrations presented here, SpectRobot provides a common framework to encode tactile sensing in learning-based robotics. Because the tactile signal is converted into an image-like representation before entering the policy, the framework is largely agnostic to the downstream learning architecture.
The same method could therefore be integrated into Vision-Language-Action models, World-Action Models, or future embodied learning architectures, provided that they contain an image-encoding stage. 
The learned representations produced by SpectRobot may also provide information beyond direct manipulation. 
Because the policy identifies signatures directly from physical interaction, these representations could help reveal which aspects of contact dynamics should be retained in future contact models. 
In this sense, high-bandwidth tactile learning could contribute to the development of more relevant models for sim-to-real transfer, where accurately describing friction, impacts, deformation, and other contact phenomena remains challenging.
As robotic control rates increase, high-bandwidth tactile sensing could increasingly support fast control, process monitoring, and predictive maintenance by enabling systems to learn directly from vibration signals to detect contact states, degradation, and incipient failure without relying solely on predefined fault frequencies or manually engineered indicators.

\section*{Acknowledgements}

We are grateful to Haixuan Xavier Tao and the organizers of the Global Open-Source Innovation Meetup (GOSIM) 2026 in Paris for the momentum and encouragement provided during the early stages of this project. 
We thank Guillaume Bonifas, Damien Frechou, and Yvic Pineau for insightful discussions on robotic manipulation, and we are grateful to Fabien Danieau for valuable exchanges on tactile sensing. 
Finally, we thank Julien Cheret and Dušan Žibrat Kalanj for sharing their expertise on Dewesoft acquisition systems and the openDAQ framework, respectively.

\section*{Conflict of Interest}

The Dragonfly\textsuperscript{\textregistered} sensors evaluated in this study are manufactured by Wormsensing, which employs two of the authors.

\section*{Data Availability}

The dataset used in this study is publicly available at: \url{https://huggingface.co/jogarulfop}.
All data are stored in the LeRobotDataset v3.0 format. 
Camera and spectrogram streams are encoded as video files.

\section*{Code Availability}

The code developed for this study is openly available under the Apache License 2.0 at: \url{https://github.com/spectrobot-project}.

\bibliographystyle{IEEEtran}
\bibliography{references}

\end{document}